# Neural Network-Based Active Learning in Multivariate Calibration

Abhisek Ukil, *Senior Member, IEEE*, and Jakob Bernasconi

*Abstract*—In chemometrics, data from infrared or near-infrared (NIR) spectroscopy are often used to identify a compound or to analyze the composition of a material. This involves the calibration of models that predict the concentration of material constituents from the measured NIR spectrum. An interesting aspect of multivariate calibration is to achieve a particular accuracy level with a minimum number of training samples, as this reduces the number of laboratory tests and thus the cost of model building. In these chemometric models, the input refers to a proper representation of the spectra and the output to the concentrations of the sample constituents. The search for a most informative new calibration sample thus has to be performed in the output space of the model, rather than in the input space as in conventional modeling problems. In this paper, we propose to solve the corresponding inversion problem by utilizing the disagreements of an ensemble of neural networks to represent the prediction error in the unexplored component space. The next calibration sample is then chosen at a composition where the individual models of the ensemble disagree most. The results obtained for a realistic chemometric calibration example show that the proposed active learning can achieve a given calibration accuracy with less training samples than random sampling.

*Index Terms*—Active learning, chemometrics, design of experiment, ensemble of models, error prediction, inverse model, model building, multivariate calibration, near infrared (NIR), spectra, spectrometer, spectroscopy.

## I. INTRODUCTION

INFRARED (IR) or near-infrared (NIR) spectroscopy is a method used to identify a compound or to analyze the composition of a material by studying the interaction of IR light with matter. Chemometrics is the application of mathematical and statistical methods to analyze chemical data, e.g., multivariate calibration, signal processing/conditioning, pattern recognition, experimental design, etc. [1].

The process of finding appropriate model parameters that lead from the spectrum to the desired information on the composition of the material is called calibration. In chemometrics, calibration is achieved by using the spectra as multivariate descriptors to predict concentrations of the constituents. The sequential steps in multivariate calibration are described in [2] and [3].

The corresponding calibration model can be linear (e.g., linear regression from standard software like GRAMS/AI [4], Unscrambler [5]), HORIZON MB [6], etc., or nonlinear. In the nonlinear calibration category, neural networks (NNs) have been widely used, as reported in the works [7]–[9], also for particular spectroscopy applications, e.g., prediction of cleaning solution component concentrations [3], identification of amino acid [10], gasoline applications [11], [12], polyethylene density prediction [13], etc.

### A. Motivation

In chemometrics, depending on the application, the chemical lab tests typically cost around 500–1000 USD per sample [3]. Therefore, it is of particular interest to achieve a given calibration accuracy with minimal number of training samples to minimize the cost of model building. For this, one often uses fixed experimental designs [14], where the entire set of training samples is determined in one step. Nonlinear calibration techniques, particularly using NN [3], [7]–[13], try to achieve a faster convergence to a particular accuracy level, basically replacing the linear methods.

### B. Problem to be Solved

In this paper, however, we shall primarily be interested in a sequential determination of optimal training samples, i.e., by active learning techniques. Our main objective will be to evaluate use of an ensemble of NN models for active learning in the calibration of chemometric models. However, due to the particular nature of the calibration in chemometrics, one cannot directly apply the state-of-the-art active learning methods. Instead, one has to solve an inverse problem for the active learning, which will be explained later. It is also interesting to verify whether such inverse modeling-based active learning method yields any tangible advantage over the standard direct methods.

The remainder of this paper is organized as follows. Section II gives an overview of different state-of-the-art active learning techniques. The inverse problem that has to be solved when active learning is applied, in particular, to chemometrics calibration problems is explained in Section III. Experimental results are described in detail in Section IV and discussed in Section V, followed by conclusions in Section VI.

## II. ACTIVE LEARNING

### A. What is Active Learning

In industrial research, one is frequently confronted with the task of exploring the relationship between a number of input variables $\mathbf{x} = (x_1, x_2, \ldots, x_n)$ and some response $y$. In

Manuscript received February 15, 2012; revised June 14, 2012; accepted September 20, 2012. Date of current version December 17, 2012. This work was supported by the Sensors and Signal Processing program, ABB Corporate Research.

The authors are with ABB Corporate Research, Baden-Dättwil CH-5405, Switzerland (e-mail: abhisek.ukil@ch.abb.com; jakob.bernasconi@alumni.ethz.ch).

chemometric calibration problems, **x** may, e.g., refer to the principal components (PCs) of a measured IR spectrum and $y$ to the concentration of a constituent in the corresponding sample.

In most cases, these relationships are so complex that it is impossible to model them reliably from physical principles, i.e., they can only be explored through a combination of experimentation and empirical modeling. The main objectives of such a process are the following:
1) to achieve an acceptable fit of the experimental data;
2) to obtain reliable information about the relationship between $y$ and **x** in a predefined $x$-region of interest;
3) to minimize the necessary number of experiments.

The first two of these points, in particular, the choice of the empirical model and of the statistical procedures are used to estimate the unknown model parameters. The third of the above objectives, the minimization of the required number of experiments, refers to the field of "experimental design" [14], i.e., to the choice of the **x**-values at which the experiments are to be made.

Most of the literature on experimental designs is concerned with fixed designs, i.e., the number of experiments to be performed is fixed in advance. The problem is then to determine at which input values the experiments should be made in order to extract as much information as possible about the relationship between $y$ and **x**.

Here, however, we are rather concerned with the problem that an experimental approach is very expensive and time consuming. If this is the case, we usually do not want to fix the number of experimental runs in advance. Sometimes, we can start from an already existing set of experimental data, and it may then be desirable or even necessary to perform the additional experiments sequentially.

Under these circumstances, we want to know "where to look next" in order to learn as much as possible about the relationship to be explored. In other words, we want to determine at which $x$-values the next experiment has to be performed so that the expected information gain is maximal. This problem of "active data selection" or "active learning" has a long history, and it has gained renewed interest in connection with NN learning [15]–[21].

In the following sections, we briefly summarize some active learning strategies that have been described in the literature. These strategies are adaptive in the sense that the decision of where to perform the next experiment depends on the outcome of all previous experiments, i.e., on our present knowledge about the relationship to be explored.

### B. Maximal Information Gain

We want to model the relationship between a vector **x** of input variables and a response $y(\mathbf{x})$ by a function $y = f(\mathbf{x}, \mathbf{w})$, where **w** denotes a vector of parameters.

The outcome of an experiment is represented as an input–response pair $\{\mathbf{x}_i, t_i\}$, where $t_i$ is the measured response at input $\mathbf{x}_i$, and $\rho_N(\mathbf{w})$ denotes the probability distribution of the model parameters **w** after $N$ measurements $\{\mathbf{x}_i, t_i\}$, $i = 1, 2, \ldots, N$, have been analyzed.

The information gain associated with making a new experiment at **x** is then equal to the expected entropy decrease

$$\triangle S(\mathbf{x}) = \int dt \rho_N(t|\mathbf{x})[S_N(\mathbf{w}) - S_{N+1}(\mathbf{w}|\{\mathbf{x}, t\})] \quad (1)$$

where

$$S_N(\mathbf{w}) = -\int d\mathbf{w} \rho_N(\mathbf{w}) \log \rho_N(\mathbf{w}) \quad (2)$$

$$S_{N+1}(\mathbf{w}|\{\mathbf{x}, t\}) = -\int d\mathbf{w} \rho_{N+1}$$
$$\times (\mathbf{w}|\{\mathbf{x}, t\}) \log \rho_{N+1}(\mathbf{w}|\{\mathbf{x}, t\}) \quad (3)$$

$$\rho_{N+1}(\mathbf{w}|\{\mathbf{x}, t\}) = \frac{\rho(t|\mathbf{x}, \mathbf{w}) \rho_N(\mathbf{w})}{\rho_N(t, \mathbf{x})}, \text{(Bayes' rule)} \quad (4)$$

and

$$\rho_N(t|\mathbf{x}) = -\int d\mathbf{w} \rho_N(\mathbf{w}) \rho(t|\mathbf{x}\mathbf{w}) \quad (5)$$

and where $\rho(t|\mathbf{x}, \mathbf{w})$ is the probability density that the model with parameter values **w** predicts a response $t$ if the measurement is made at input **x**. According to the strategy of maximal information gain, the next experiment should then be made at that **x**-value for which the expected entropy decrease $\triangle S(\mathbf{x})$ is maximal.

### C. Maximum Modeling Uncertainty

If we assume that the empirical model $y = f(\mathbf{x}, \mathbf{w})$ is capable of representing the experimental data $\{\mathbf{x}_i, t_i\}$ up to a zero-mean Gaussian noise with standard deviation $\sigma$, an approximate evaluation of $\triangle S(\mathbf{x})$ leads to the following expression [16]:

$$\triangle S(\mathbf{x}) = \frac{1}{2} \log[1 + V(\mathbf{x}, \mathbf{w}_N)/\sigma^2] \quad (6)$$

where $\mathbf{w}_N$ refers to the parameter values estimated from the first $N$ experiments, and $V(\mathbf{x}, \mathbf{w}_N)$ measures the corresponding modeling uncertainty. This result tells us that we obtain maximal information gain if we make our next experiment at the **x**-value with the largest modeling uncertainty.

For nonlinear models, e.g., NN, the evaluation of $V(\mathbf{x}, \mathbf{w}_N)$ is computationally quite expensive and can create numerical problems. The modeling uncertainty can, however, be estimated empirically by the variance of a set of randomly generated model approximations, and there exist several straightforward procedures for a suitable generation of different model approximations.

1) One possibility is to add Gaussian noise with variance $\sigma^2$, to the existing experimental data, creating different sets of pseudomeasurements with which the empirical model can be trained ($\sigma$ refers to the known or estimated experimental error).
2) In the case of NNs, there exist a number of additional methods to create different model approximations from a given set of experimental data. We can, e.g., start the training process from different, randomly chosen, initial weights **w**, or use bootstrapping [3] to generate different sets of training samples, etc.

## D. Space-Filling Strategies

A completely different approach to active learning is space-filling strategy. In space-filling, the next experiment is simply performed at the **x**-value that has the largest "distance" $d(\mathbf{x})$ to the existing training data. As a measure of distance, one can usually choose the Euclidean distance in **x**-space

$$d(\mathbf{x}) = \min_k \| \mathbf{x} - \mathbf{x}_k \| \qquad (7)$$

where the minimum is taken over the current set of training samples. This will be discussed in detail later with respect to calibration problems.

## E. Discussion and Comparison of Active Learning Strategies

The performance of active learning strategies has been tested on a number of (mostly academic) problems [15]–[21]. Hu *et al.* [17] proposed unsupervised active learning using graph-theoretic approach. Xingquan *et al.* [18] used ensemble of different classifiers for active learning from streaming data. Basak and Gupta [19] used feedforward NN for active learning. Krogh and Vedelsby [20] compared NN ensembles for active learning. Poland and Zell [21] have used five simple test functions to compare the performance of different active learning strategies with that of random data selection. In their tests, the variance-based methods (maximal modeling uncertainty) always led to a considerably faster decrease of the modeling error (with increasing number of training samples) than random or space-filling strategies. Only for low-dimensional input spaces, space-filling led to a significant improvement over random strategies.

Space-filling strategies, on the other hand, are simpler to implement and, in general, more robust than strategies based on information gain or modeling uncertainty. Another problem with the variance-based active learning is that these strategies rely on the assumption that the model used is capable of representing the (unknown) relationship between the input **x** and the response $y$ (up to measurement noise). If the approximation model is too simple, the modeling uncertainty may give a misleading answer to the question of where to choose the next experiment. A nonuniform measurement uncertainty, e.g., much higher measurement errors in some regions of the input space, can also lead to problems with variance-based strategies.

## III. ACTIVE LEARNING IN CALIBRATION PROBLEMS

### A. Inverse Problem

All of the active learning strategies discussed in the previous section refer to a learning problem where the training samples are constructed by performing an experiment (i.e., measuring the output $y$) for a chosen value of the input **x**. In this paper, however, we are concerned with a different type of learning problems. These refer to the calibration of chemometric models [1], [3] that predict, e.g., the concentration of some constituent (output $y$) from a condensed representation (input **x**) of the IR spectrum (PCs, partial least squares [1], or wavelet components) of a heterogeneous material. The search for a most informative next experiment thus has to be performed in the output space, rather than in the input space as in conventional modeling problems. As one cannot measure the "next spectrum," one has to choose the sample constituents to get the next composition, the spectrum of which is then measured.

For space-filling strategies, a corresponding adaptation is straightforward. The space-filling selection of new training samples is simply performed in the output space, i.e., in the space of concentrations and other measurement characteristics. Space-filling methods can thus be used both for the construction of an optimal initial training set and for the sequential addition of new training samples [14].

The adaptation of variance-based active learning strategies for calibration, on the other hand, leads to a number of challenging problems. In particular, we have to solve an inversion problem, i.e., to transform the information about modeling uncertainties from the input (spectra) to the output (concentration) space. We can, however, not simply sample the input space (e.g., uniformly) for data points with high uncertainty and then transfer this information to the output space. The reason is that a given region in the concentration space, e.g., is mapped in an unknown and perhaps very nonuniform way into the space of PCs.

One possibility to approach the inversion problem is to measure the modeling uncertainty for the training samples, for which the concentrations are known, and then use this information to train an "inversion model" that predicts an estimation of the modeling uncertainty for all points in concentration space.

### B. Algorithm

The inverse problem of active learning in calibration problems is described in the foregoing section. In the following, we describe our proposed algorithm to solve this inversion problem. The models used in the algorithm all refer to feedforward NNs [22], [23] that are trained by error backpropagation [22], [23].

For simplicity, we formulate our active learning algorithm for models that predict the concentration of a single component. The variance-based active learning algorithm is described as follows.

1) *Step 0:* Choose an initial set of $N_0$ calibration samples. Each sample consists of the measured concentration and of the corresponding IR spectrum (characterized, e.g., by a number of PCs).
2) *Step 1:* Train an ensemble of "prediction models" (models that predict the concentration of a sample from the measured IR spectrum) with the current set of calibration samples. Bootstrapping [3] is used to create the individual models of the ensemble.
3) *Step 2:* Determine the concentration prediction (= average over the individual predictions) and the "modeling uncertainty" (= variance of the predictions of the individual models in the ensemble) for all calibration samples.
4) *Step 3:* Train an "inversion model" (prediction of the modeling uncertainty for a given concentration) with the corresponding data for the current set of calibration samples.

5) *Step 4:* With the "inversion model," predict the modeling uncertainty for an appropriately chosen set of concentration values.
6) *Step 5:* From this set, choose the $n_0$ concentration values (e.g., $n_0 = 1, 5$, or 20), where the predicted "modeling uncertainty" is highest. Then, prepare corresponding samples and measure their IR spectra. Add these samples to the set of calibration samples and go back to Step 1.

Iterate the algorithm until the number of calibration samples is sufficient to achieve an acceptable accuracy of the concentration prediction.

If the concentrations of several components have to be predicted, the above active learning algorithm can be applied separately for each component. Alternatively, one can define an overall "modeling uncertainty" that adequately combines the modeling uncertainties for the different concentrations.

## IV. Numerical Results

### A. Dataset

To validate the proposed active learning algorithm of Section III-B, we use the "Tecator" dataset [24]. The dataset consists of 240 NIR-spectra of several cuts of meat, along with parameters like fat, protein, and water content of each meat sample. We used all the 240 spectra and selected the fat concentration as the parameter of interest. We noticed that there are some repeated data, with identical spectra and concentration values. Nevertheless, we included all data in the models.

### B. Models

The NNs used in the different steps of the active learning algorithm are described below. The MATLAB NN toolbox [25] was used for the network design and for the experiments.

1) The "prediction networks" used in Step 1 (models to predict the fat value of the training samples) has a 10-7-3-1 architecture. That is, it has an input layer of ten nodes (ten PCs of the spectra are used as input), two hidden layer, consisting of seven and three neurons, respectively, and one output (fat content). The nonlinear transfer function between the input and the first hidden layer is a tangent-sigmoid function [22], [23], [25], and a linear transfer function is used between the first and the second hidden layer and between the second hidden layer and the output. These network design parameters have been optimized in a series of experiments. In order to avoid overfitting [22], [23], an early stopping criterion is used, each network being trained for 100 epochs.
2) The network used as "inversion model" in Step 3 (prediction of modeling uncertainty) has a simpler architecture of 1-2-1, in order to avoid overfitting. That is, it has an input layer with one input (fat content), one hidden layer with two neurons, and one output for the modeling uncertainty (standard deviation of ensemble of prediction models). The transfer function between the input and the hidden layer is tangent-sigmoid and that between the hidden and the output layer is linear. This network was trained for 20 epochs.
3) We note that to predict the fat content (prediction models), a relatively complex network structure is used. This is because we need a good accuracy in predicting the values out of which the prediction uncertainty is estimated. However, to predict the modeling uncertainty (inversion model), a relatively simple network is used. If the inversion model is chosen too complex, it would model perfectly for the training set, i.e., for data for which we have the true values, but it would generalize badly to data with unknown modeling uncertainty, which is the really important purpose of the model. Therefore, the structure of the "inversion model" is kept simple intentionally so that it might model not too well for the training set, but would predict generally well the modeling errors in the unexplored part of the component space, where we do not have the true values.

### C. Experiments

With the aforementioned network structures, the following experiments have been performed.

1) Out of the 240 samples, a fixed set of 40 samples is used as a validation set which is never used in the training. The validation set is used to see whether active learning is providing any advantage over, say, random sample addition. Therefore, after each run of choosing the next sample(s) from our active learning principle, the same number of samples is chosen randomly. For each new training set, both for active and for random sample addition, prediction on the validation set is done to compare the performances. One experiment is performed with space-filling and active learning to choose the next samples.
2) Different experiments are performed with different sizes of the starting training set (with 20, 40, and 80 starting samples out of the available 200 samples). That is, for 20 starting samples, e.g., 20 samples are randomly chosen out of the 200 existing samples, and the remaining 180 samples are kept as the buffer, from which the next samples are chosen.
3) For each number of starting samples, five different randomly chosen sets are used, and the results represent corresponding averages.
4) Different experiments are performed in terms of how many samples are added in each active learning step (5 or 20 samples per iteration). The predicted ensemble standard deviations are sorted in descending manner and the 5 or 20 values with the highest model disagreements are chosen.
5) Different experiments are performed with different ensemble sizes. Bootstrapping [3] is used for the construction of the ensembles, and the different networks in the ensemble have different initializations.
6) For comparison on the validation dataset, the root mean square error (RMSE) of the different artificial NN predictions is chosen, where RMSE is defined as

$$e_{\rm rms} = \sqrt{\frac{\sum(y - y_d)^2}{n}} \qquad (8)$$

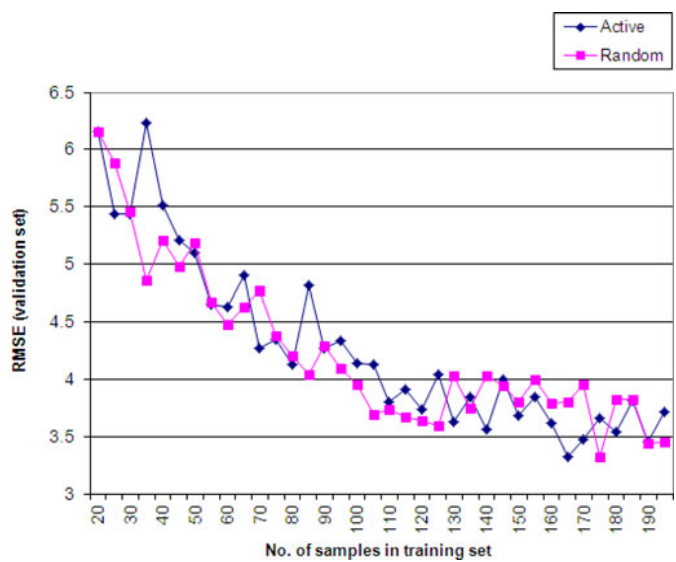

Fig. 1. RMSE of fat content prediction on the validation set for active learning and random sample addition. Addition of five samples per iteration, starting with 20 initial training samples.

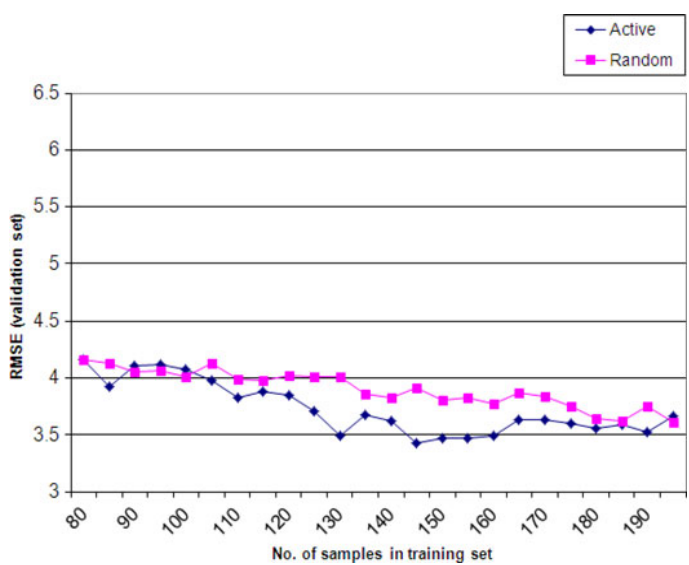

Fig. 3. RMSE of fat content prediction on the validation set for active learning and random sample addition. Addition of five samples per iteration, starting with 80 initial training samples.

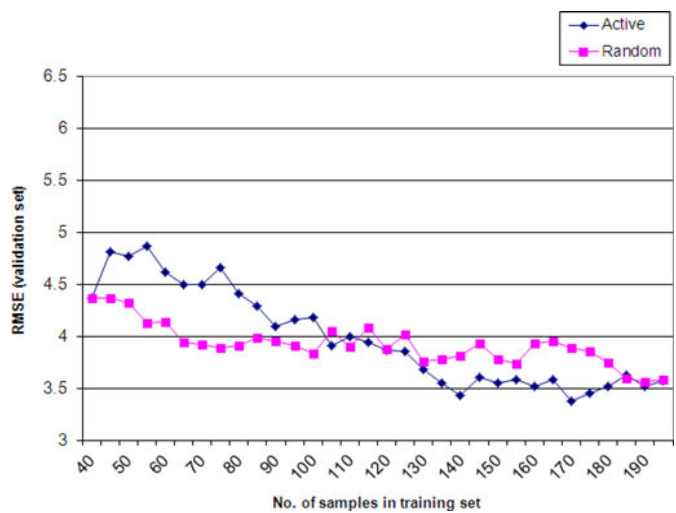

Fig. 2. RMSE of fat content prediction on the validation set for active learning and random sample addition. Addition of five samples per iteration, starting with 40 initial training samples.

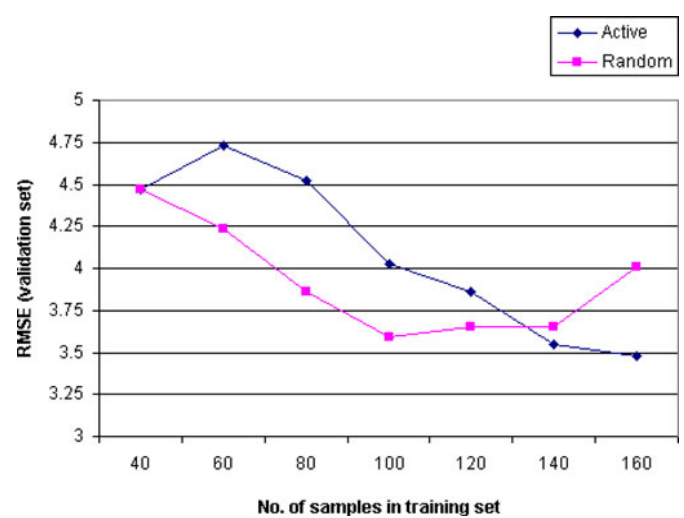

Fig. 4. RMSE of fat content prediction on the validation set for active learning and random sample addition. Addition of 20 samples per iteration, starting with 40 initial training samples.

where $y$ is the calculated output of the network, $y_d$ is the desired output, and $n$ is the number of validation samples.

*D. Results*

Figs. 1–3 show a comparison of the RMSE results on the validation set for data addition via active learning and random sampling, for an addition of five samples per iteration, starting with 20, 40, and 80 initial training samples, respectively. After several experiments, the optimal size of bootstrap models turned out to be 100 networks.

From Figs. 1–3, we see that at the end of the sample additions, i.e., at a training set size of about 195 samples, all the three experiments achieve similar RMSE levels. Minor differences are due to reasons like the difference in datasets, randomization in NN training.

Figs. 4 and 5 show a comparison of the RMSE results on the validation set for data addition via active learning and random sampling, for an addition of 20 samples per iteration, starting with 40 and 80 initial training samples, respectively.

Let us consider a target RMSE of 3.5 for the validation set, with five samples addition per iteration. From Fig. 1, we can see that with 20 starting samples, it is achieved by active learning with about 160 samples, and by random sampling with about 175 samples. From Fig. 2, we can see that with 40 starting samples, it is achieved by active learning with about 140 samples, and by random sampling with about 180 samples. From Fig. 3, we can see that with 80 starting samples, it is achieved by active learning with about 145 samples, and by random sampling with about 180 samples.

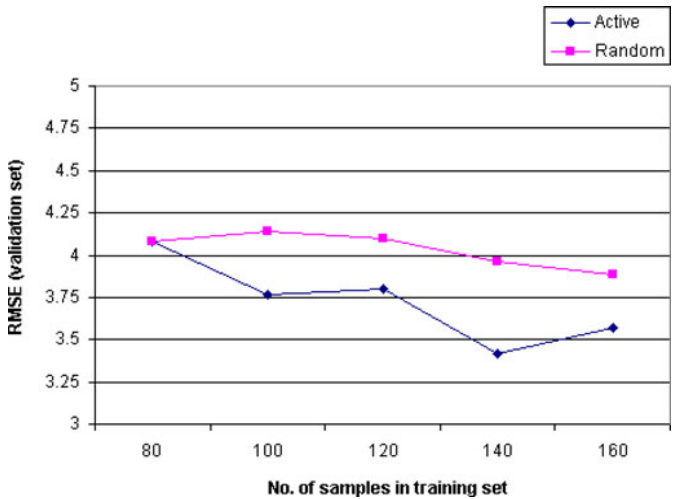

Fig. 5. RMSE of fat content prediction on the validation set for active learning and random sample addition. Addition of 20 samples per iteration, starting with 80 initial training samples.

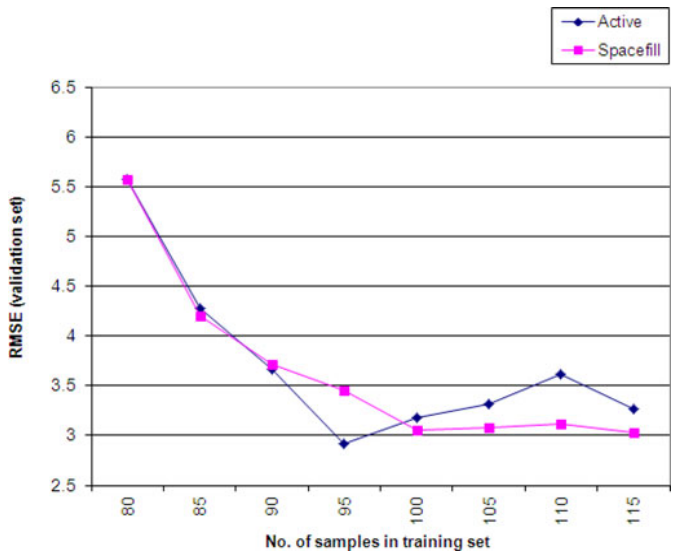

Fig. 6. RMSE of fat content prediction on the validation set for active learning and space-filling. Addition of 20 samples per iteration, starting with 80 initial training samples.

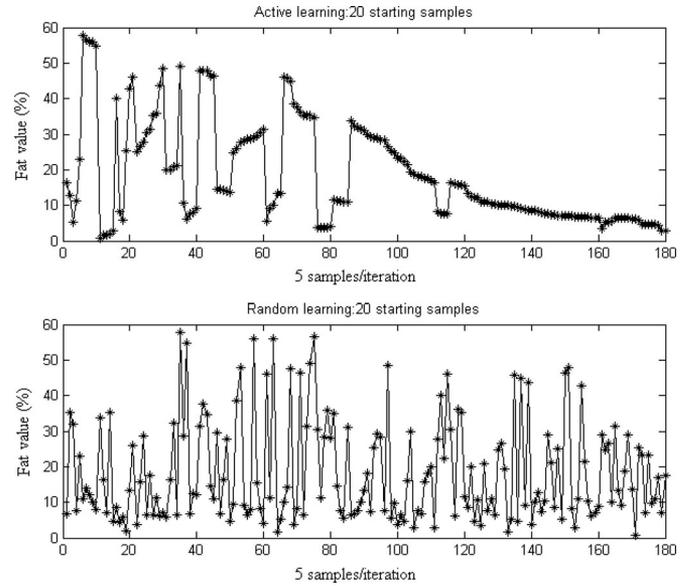

Fig. 7. Data addition via active learning and random sampling. Addition of five samples per iteration, starting with 20 initial training samples.

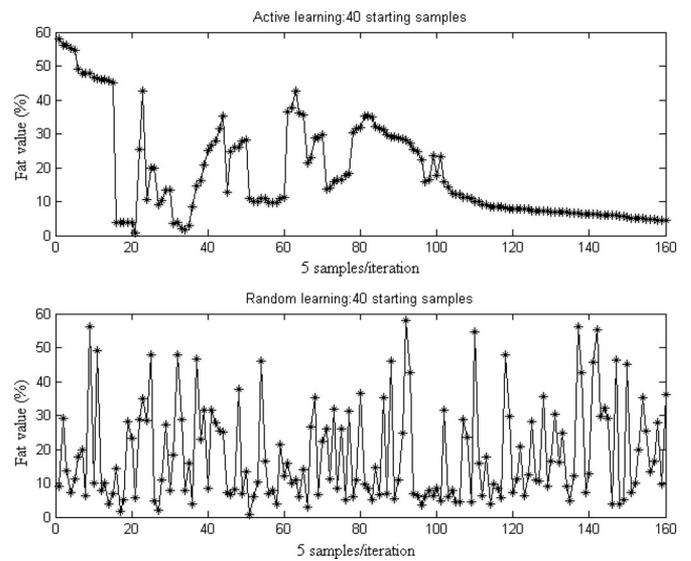

Fig. 8. Data addition via active learning and random sampling. Addition of five samples per iteration, starting with 40 initial training samples.

For the target RMSE of 3.5 for the validation set, with 20 samples addition per iteration, from Figs. 4 and 5, we see that it is achieved by active learning with about 150 and 140 samples, respectively, while it is never quite achieved with random sampling.

Therefore, we can see that with active learning, a good accuracy is obtained with a smaller number of training samples than with random sample addition.

Fig. 6 shows a comparison of the RMSE results on the validation set for data addition via active learning and space-filling, for an addition of five samples per iteration, starting with 80 initial training samples. For space-filling, the total data range (% of fat content) was divided into ten equispaced regions, and the initial samples were chosen to cover the ranges uniformly. The same starting dataset was used for the active learning. As the results converged to required accuracy level, the learning was continued till about 120 samples. From Fig. 6, we can notice that the target RMSE of 3.5 can be achieved via active learning with about 90 samples, while space-filling requires about 95 samples.

### E. Comparison of Sample Trend

We further kept a track of the added samples in order to observe the trend. Comparative data addition via active learning and random sampling are shown in Figs. 7–9 (corresponding to Figs. 1–3) for an addition of five samples per iteration, starting with 20, 40, and 80 initial training samples, respectively, and in Figs. 10 and 11 (corresponding to Figs. 4 and 5), for an addition of 20 samples per iteration, starting with 40 and 80 initial training samples, respectively. Fig. 12 (corresponding to Fig. 6) shows comparative data addition via active learning and space-filling for an addition of five samples per iteration, starting with 80 initial training samples. In Figs. 7–12, the *Y*-axis represents the fat value as a percentage.

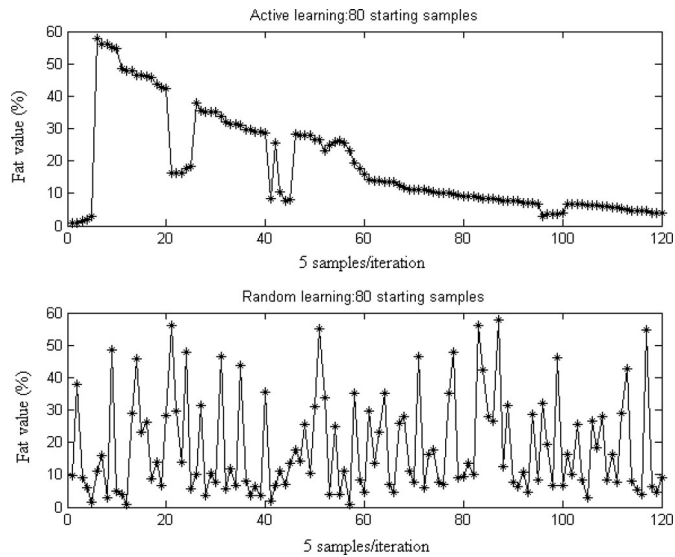

Fig. 9. Data addition via active learning and random sampling. Addition of five samples per iteration, starting with 80 initial training samples.

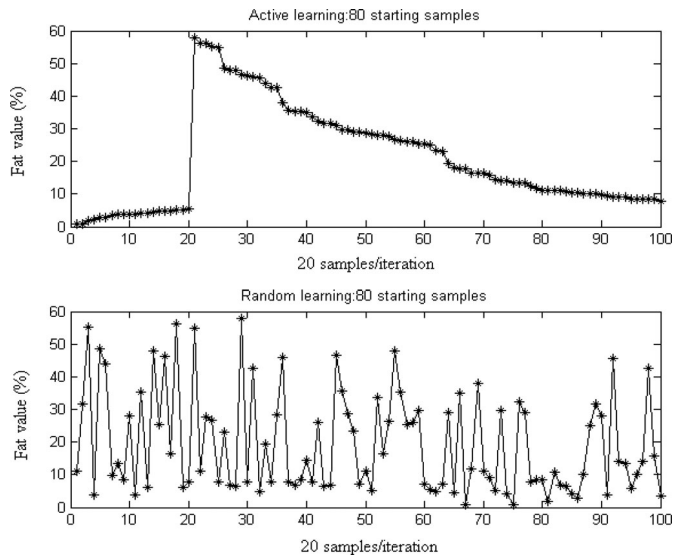

Fig. 11. Data addition via active learning and random sampling. Addition of 20 samples per iteration, starting with 80 initial training samples.

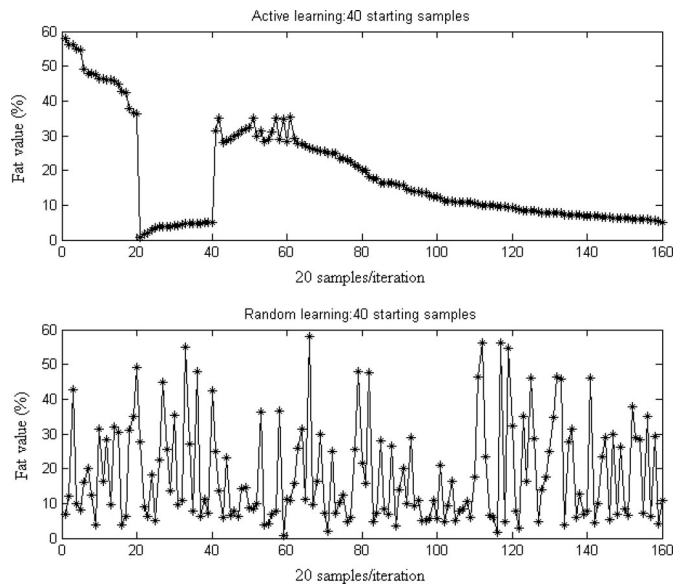

Fig. 10. Data addition via active learning and random sampling. Addition of 20 samples per iteration, starting with 40 initial training samples.

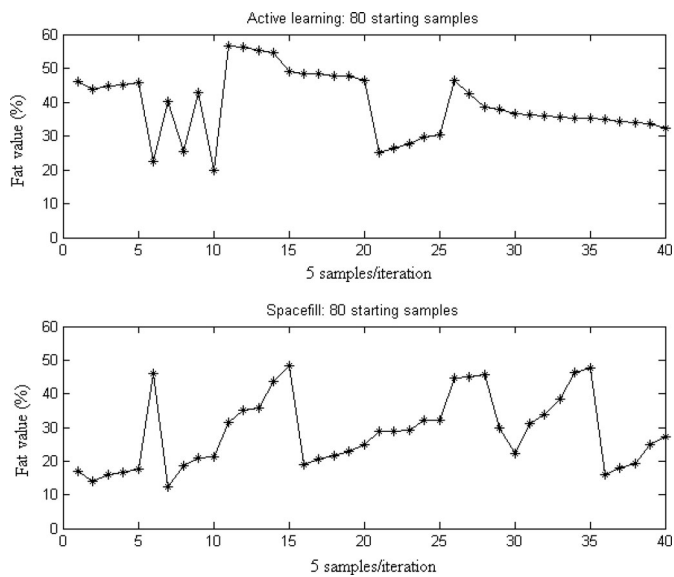

Fig. 12. Data addition via active learning and space-filling. Addition of five samples per iteration, starting with 80 initial training samples.

## V. Discussion

With respect to the above results, we would like to add the following comments.
1) An interesting aspect for data addition is to achieve a particular accuracy level (in our case, over the validation dataset) with a minimum number of training samples, as this reduces the number of laboratory tests and thus the cost of model building [3]. From Fig. 1, we see that starting with 20 initial samples, the active learning is not much better than random sample addition. From Fig. 2, one can see that active learning performs even worse than random sample addition up to about 110 training samples. After that, however, it converges to a particular accuracy level faster than random addition of samples. The effect of active learning is even more prominent in Fig. 3 for the case of 80 starting samples. Overall, active learning performs well and converges faster to a particular accuracy level (e.g., in our case target RMSE of 3.5 for validation set) than random sample addition.

2) We notice that when we start with a low number of initial training samples, e.g., 20, there is not much gain with active learning over random sampling in the initial part. This is probably due to the fact that active learning when employed from the very beginning without a considerable knowledge about the system might drive the system in a very special direction, e.g., concentrating on difficult samples, and thereby failing to achieve a good overall generalization ability for the validation samples. However, if we first choose a larger amount of initial samples and then employ our model-inversion-based active learning, we notice a consistently better performance, getting down to a particular accuracy level significantly faster than random sampling. Therefore, the proposition would be to start with a not too small set of initial samples before applying active learning for better convergence. For the "Tecator" dataset, we observed an optimal starting set of about 40–80 samples. The lower this number, the better is active learning for the overall cost of model building.
3) For an addition of 20 samples per iteration, we also observe a better convergence capability of active learning over random sampling (see Figs. 4 and 5). There, we also notice that starting with 80 initial training samples is more advantageous for active learning than starting with a lower number of samples.
4) From the nature of the added samples (see Figs. 7–11), we can see that in contrast with random sampling, active learning in general shows a specific systematic pattern in choosing the samples. Note that we perform the inverse modeling based on the highest discrepancies in the ensemble of the prediction models. Therefore, active learning tries to systematically cover the ranges where the models disagree most, i.e., where they are less certain. For the "Tecator" set, it seems that the prediction of the higher fat values is more difficult, which is seen in Figs. 7–9, where the active learning process tries to cover the ranges from the higher to lower fat values.
5) From the nature of the added samples via active learning and space-filling (see Fig. 12), we can notice that space-filling tries to fill data uniformly (in the range of 0–60% fat content) in each iteration of sample addition. The active learning, as mentioned before, tries to cover the ranges from the higher to lower fat values, which is different than straight forward space-filling. As discussed before and from Figs. 1–6, we can see that active learning is quite advantageous over random sample addition, and bit more advantageous than space-filling in this example case.
6) From Fig. 6, we can see that if the initial set is chosen via space-filling instead of random sampling (see Figs. 1–5), the target accuracy can be achieved faster (e.g., in this case with about 90 samples).
7) Due to the particular nature of the calibration in chemometrics, one cannot directly apply the state-of-the-art active learning methods [15]–[21]. As one cannot measure the "next spectrum," one has to choose the sample constituents to get the next composition, the spectrum of which is then measured. Therefore, one has to solve an inverse problem for the active learning, as demonstrated in this paper. Nevertheless, this is very specific to the spectroscopy and chemometrics field.
8) Comparing Figs. 7–9, one can see that even with active learning, there are some variations (automatically invoked by the active learning scheme) in choosing the samples. For example, in Fig. 7, for 20 starting samples, the first 85 chosen samples show quite some fluctuations in their values, while these fluctuations are much less pronounced for 80 starting samples (see Fig. 9). This probably indicates that 80 starting samples is an optimal number for the "Tecator" dataset.
9) For an addition of 20 samples per iteration (see Figs. 10 and 11), the variation in the values of the added samples is smaller. For example, the first 20 samples added in Fig. 11 (one iteration with 20 samples) all have a small fat content. In Fig. 9 (four iterations with five samples per iteration), for comparison, only the first five samples have a low fat content, while the next 15 samples have a high fat content. It is to be noted that in both cases, the initial 80 training samples are identical.
10) In our example of the "Tecator" dataset, we had a total of 200 preacquired samples. Then, we started with 80 samples, and kept 120 in buffer from which we chose actively. In practice, this would be analogous to asking the experimenter to make the next samples there. For example, in Fig. 9, starting with 80 initial samples, for the next round's five sample addition, one would ask the experimenter to choose low fat samples, e.g., in the range of 0–5% (value along the $Y$-axis). This would be the idea of the active learning.
11) From Fig. 3, if one assumes a realistic target RMSE of 3.5 for the given application, one can achieve this with about 145 samples with active learning, while one would require about 180 samples using random sampling. This means a saving of 35 samples. In chemometrics, the cost of model building depends on the number calibration samples, as one has to perform lab tests to determine the actual values. Depending on the application, the typical costs of chemical lab tests could be in the range of 500–1000 USD/sample [3]. Therefore, active learning can provide a significant advantage in cost-effective model building.
12) Our active learning algorithm, which uses the disagreements in the ensemble of model predictions, proposes the next optimal samples automatically by solving the inverse modeling problem in chemometrics. This does not require any particular system background knowledge, as is typically required to design the experiments [14]. However, in order to optimally use the proposed active learning method, one does require a realistic number of initial samples; therefore, one actually uses a considerable amount of system knowledge to estimate the uncertainties in the remaining component space. For the "Tecator" dataset, our experiments indicate that the optimal initial starting set size is about 80 samples.
13) As demonstrated, one can keep the number of samples to be added as a parameter. In principle, one can set it to one

sample per iteration. However, we skipped this choice as it would have taken a five times higher computation time than an addition of five samples per iteration. In addition, in chemometrics, one typically is not interested in a single next sample, but rather a bunch of samples to utilize the chemical lab tests in a cost-optimal way. Five samples seem to be an acceptable value for a sample bunch to be added.

## VI. Conclusion

In a multivariate calibration process, one finds the optimal model parameters that lead from the spectrum to the desired information on the composition of the material. An interesting aspect for data addition is to achieve a particular accuracy level with a minimum number of calibration samples, as this reduces the number of laboratory tests and thus the cost of model building. In chemometric calibration models, the input refers to a proper representation of the spectrum and the output to the concentrations of the sample constituents. The search for a most informative next experiment thus has to be performed in the output space, rather than in the input space as in conventional modeling problems. In this paper, we have proposed an inverse modeling technique that uses the disagreements of an ensemble of prediction models to determine the next samples in the unexplored component space. We propose to choose the next sample(s) where the models disagree most, i.e., calibration samples for which the prediction models are most uncertain. This algorithm was tested using the "Tecator" dataset [24]. The comparative results show that the proposed active learning algorithm can be beneficial in terms of attaining a particular accuracy level with a smaller number of samples than for the case of random sample addition. It is also observed that the better convergence using the active learning algorithm can only be obtained if we have accumulated a minimum amount of system knowledge, i.e., if we use a minimal amount of initial calibration samples.

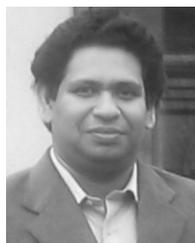

**Abhisek Ukil** (S'05–M'06–SM'10) received the Bachelor of electrical engineering degree from the Jadavpur University, Kolkata, India, in 2000, and the M.Sc. degree in electronic systems and engineering management from the University of Bolton, Bolton, U.K., in 2004. He received the Ph.D. degree from the Pretoria (Tshwane) University of Technology, Pretoria, South Africa, in 2006, working on power systems disturbance analysis with Eskom.

From 2000 to 2002, he was a Software Engineer with InterraIT, Noida, India. In 2006, he joined the Integrated Sensor Systems Group, ABB Corporate Research, Baden-Dättwil, Switzerland, where he is currently a Principal Scientist. He is an author/coauthor of more than 45 refereed papers, a monograph, two book chapters, and inventor/co-inventor of seven patents. His research interests include signal processing, machine learning, power systems, embedded systems, and sensors.

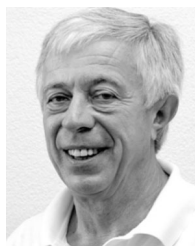

**Jakob Bernasconi** received the Diploma and Ph.D. degrees in theoretical physics from the ETH Zurich, Zurich, Switzerland, in 1968 and 1972, respectively.

He was a Research Fellow with the ABB Corporate Research, Baden-Dättwil, Switzerland, where he is currently a consultant. He has contributed to more than 80 scientific papers. His research interests include disordered systems, stochastic processes, neural networks, machine learning, and economic modeling.